\pdfoutput=1
\documentclass[11pt,a4paper]{article}
\usepackage[table]{xcolor}
\usepackage[hyperref]{emnlp2020}
\usepackage[utf8]{inputenc}
\usepackage[T1]{fontenc}
\usepackage{times}
\usepackage{latexsym}
\usepackage{graphicx}
\usepackage{amsmath}
\usepackage{booktabs}
\usepackage{multirow}
\usepackage{adjustbox}

\usepackage{microtype}

\aclfinalcopy 
\def\aclpaperid{SIGTYP 4}

\definecolor{light-gray}{gray}{0.95} 

\newcommand{\splitcond}[1]{%
  \begin{tabular}[t]{@{}l@{}}#1\end{tabular}%
}

\title{NEMO: Frequentist Inference Approach to Constrained Linguistic
  Typology Feature Prediction in SIGTYP 2020 Shared Task}

\author{Alexander Gutkin \\
  Google, London, United Kingdom \\
  \texttt{agutkin@google.com} \\\And
  Richard Sproat \\
  Google, Tokyo, Japan \\
  \texttt{rws@google.com} \\}

\date{}

\begin{document}
\maketitle

\begin{abstract}
  This paper describes the NEMO submission to SIGTYP 2020 shared task
  \cite{st-overview2020sigtyp} which deals with prediction of linguistic typological features for
  multiple languages using the data derived from World Atlas of
  Language Structures (WALS). We employ frequentist inference to
  represent correlations between typological features and use this
  representation to train simple multi-class estimators that predict
  individual features. We describe two submitted ridge
  regression-based configurations which ranked second and third
  overall in the constrained task. Our best configuration achieved the
  micro-averaged accuracy score of $0.66$ on 149 test languages.
\end{abstract}

\section{Introduction}
The rapidly developing field of computational linguistic
typology~\cite{ramat2011} is becoming increasingly popular in natural language
processing (NLP) research~\cite{bender2016,ponti2019}, where typological corpora
such as the World Atlas of Language Structures (WALS)~\cite{wals} and
\textsc{autotyp}~\cite{autotyp} are seeing increased
use~\cite{naseem-etal-2012-selective,burdick2020}. Despite their popularity,
typological corpora are very sparse. According to~\citet{murawaki-le-2018}, less
than 15\% of the feature values are present for the 2,679 languages represented
in WALS. These databases are human-curated and depend on grammatical
descriptions for their development; sparsity is often due to linguistic sources
lacking data on particular features for a given language. Developing
methodologies for accurately predicting missing typological features on the
basis of existing knowledge is therefore crucial for a wider adoption of
typological resources in NLP tasks and beyond~\cite{evans2009}.

This paper presents the work done by the ``NEMO Team'' (Google London
and Tokyo) on the \emph{constrained subtask} for the \textsc{sigtyp}
2020 Shared Task \cite{st-overview2020sigtyp}. We experimented with a variety of machine learning
models using only the features provided in the training, development
and test sets. Our features included genetic features (genus and
family), areal features (clusters of languages within a radius of the
target language), and derived implicational universals. Originally
introduced by~\citet{greenberg1963} and demonstrated to capture the
syntactic typology well~\cite{dryer1992,dryer2009,dunn2011}, similarly
to others~\cite{daume2007} we use the framework of universal
implications to capture correlations between other types of
typological features as well. We describe how these features were
derived in detail in Section~\ref{sec:method}.

As we report below in Section~\ref{sec:experiments}, the performance
of machine learning algorithms varied across different feature
predictions, with some algorithms working better for some features,
and less well for others. On balance however we found that ridge
regression~\cite{hoerl1970}, also known as Tikhonov
regularization~\cite{franklin1974}, was the most useful approach.

\section{Related Work}
\label{sec:related}
Here we review the approaches corresponding to the constrained sub-task
of the shared task, where no external data, such as unlabeled
texts, is allowed.

A popular approach to modeling the typological diversity of the world's
languages is based on Bayesian probabilistic inference. Despite recently
drawing some criticism on linguistic grounds by~\citet{ono2020}, this
approach possesses impressive explanatory power.  In what possibly
represents the earliest model-based typological feature
imputer, \citet{daume2007} introduced the probabilistic Bayesian model
for uncovering universal implications in WALS data by associating
random variables with individual WALS features and discovering the
inter-feature correlations from statistical dependence between random
variables. The modeling power of the Bayesian approach was further
demonstrated by~\citet{daume-iii-2009-non} in a non-parametric
hierarchical Bayesian model combining linguistic areas and phylogeny.

\citet{murawaki-2015-continuous} proposed a deep learning approach to
phylogenetic inference by mapping the language vectors to a latent
space using an auto-encoder trained using typologically-inspired
objective on WALS data with missing values imputed using a regularized
iterative variant of Multiple Correspondence Analysis
(MCA)~\cite{josse2012,josse2012mca}. In later
work, \citet{murawaki-2017-diachrony,murawaki2019bayesian}
and \citet{murawaki-le-2018} abandoned an earlier model in favor of a
Bayesian autologistic approach and demonstrated the superiority of
Bayesian predictor over MCA. In this approach, the languages are
represented as random variables that are explained in terms of other
languages related to each other through phylogenetic and spatial
neighborhood graphs. \citet{bjerva-etal-2019-probabilistic} introduce
a generative model inspired by the Chomskyan
principles-and-parameters framework, drawing on the correlations
between typological features of languages to tackle the novel task of
typological collaborative filtering, a concept borrowed from the area
of recommender systems.

While most state-of-the-art missing feature imputation methods are
model-based, recently \citet{buis2019} employed the iterative
technique based on singular value decomposition (SVD) from the
well-studied area of low-rank matrix completion and reported the
performance on par with the prior art. Although lacking explanatory
power, similarly to MCA, such techniques are attractive due to their
simplicity and computational efficiency.
\citet{takamura2016} took a standard machine learning approach by
training multinomial logistic regression classifiers for individual
WALS features based on other features present in the database under
various experimental conditions, hypothesizing that the classifier would
capture feature correlations implicit in the data.

The frequentist approach we take is similar in some respects to work
of~\citet{takamura2016} in that we also train vector space
classifiers, but there are a few notable differences, we: (i) use
typologically-motivated ``probabilistic'' frequency-based input space
by explicitly representing areal and phylogenetic associations and
implicational universals, and (ii) explore a wider range of classification
approaches.

Although for this task there is a special interest in the geographic
aspect of the modeling, for our final submission we limited ourselves
to the rather orthodox approach of representing language associations
through fixed neighborhoods, in order not to over-complicate our
method (described in the next section) that can be difficult to
reconcile with the more sophisticated models, such as the very
promising model of language evolution from~\citet{kauhanen2019} and
the findings emerging from the fields of dialectology and
dialectometry~\cite{szmrecsanyi2011,wieling2015,nerbonne2020}.

\section{Method}
\label{sec:method}
Here we outline the details of our approach used to generate the final
submission. The open-source implementation of our training and
evaluation pipeline has been released in public domain.\footnote{Available
  at
  \url{https://github.com/google-research/google-research/tree/master/constrained\_language\_typology}.}

\subsection{Precomputation of Features}
\label{subsec:precomp}
Typological features were preprocessed to find likely associations between genetic and areal
properties of the language. For each typological feature $f$ and value $v$ from a set of its
values $V_f$ we computed the following probability estimates:
\begin{align}
p_{\mathrm{genus}}(v|f) &= \frac{\mathrm{count}_{\mathrm{genus}}(v)}{\mathrm{count}_{\mathrm{genus}}(f)}\label{eqn:genus}\\
p_{\mathrm{family}}(v|f) &= \frac{\mathrm{count}_{\mathrm{family}}(v)}{\mathrm{count}_{\mathrm{family}}(f)}\label{eqn:family}\\
p_{\mathrm{area}}(v|f) &= \frac{\mathrm{count}_{\mathrm{area}}(v)}{\mathrm{count}_{\mathrm{area}}(f)}\label{eqn:area} \;\; ,
\end{align}
where $\mathrm{count}(f)=\sum_{v_i \in V_f}\mathrm{count}(v_i)$. Here,
``family'' and ``genus'' in equations~\ref{eqn:genus}
and~\ref{eqn:family} were as given in the data, and ``area'' in
equation~\ref{eqn:area} comprised all languages within a 2,500
kilometer radius around the target language's latitude and longitude computed using the Haversine
formula~\cite{robusto1957}.\footnote{A reviewer noted that the hard limit of 2,500 kilometers seemed 
arbitrary and wondered why we do not weight ``neighboring languages according to closeness and use weighted clustering instead.'' We agree that in principle more sophisticated approaches would be nice, but one should 
bear in mind that the geographic centroids for languages provided in the data are at best crude, and so
doing anything more sophisticated seemed to us to be crude. Also, to do this properly, distance is really
not sufficient: one would also need to account for the presence of possible barriers to contact, including
impassable mountain ranges, seas, and hostile neighbors, elements that would be hard to model.}

In addition we computed a set of \emph{implicational universals}
\cite{greenberg1963}. For each feature value $v_j$, for feature $f_j$, we
compute the probability of of $v_j$ given $f_j$, and each ${f_i},{v_i}$ pair from the
set of known feature-value pairs in the data
\begin{equation}
p(v_j|f_j,f_i,v_i) = \frac{\mathrm{count}(v_j)}{\mathrm{count}(f_j,f_i,v_i)} \, .\label{eqn:impl}
\end{equation}

\begin{table}
  \centering
  \includegraphics[width=\linewidth]{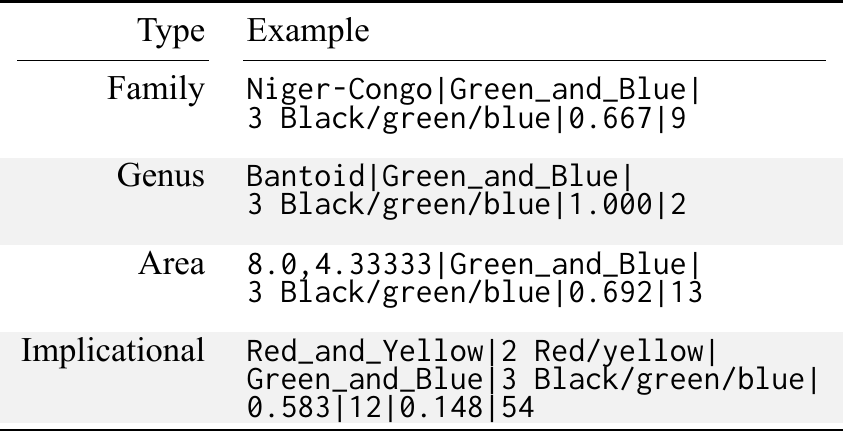}
  \caption{Examples of precomputed most likely associations for color features.}
  \label{tab:association-examples}
\end{table}

For each of the genetic (family and genus), areal (neighborhood) and
universal implication types of associations we kept a separate table, where,
for all the known features $f$ we stored
\begin{itemize}
\item the feature $f$,
\item the total number of samples in the data of the given type with
  $f$, denoted $c(f)$,
\item the value $v$ with the highest estimated probability per the
  above equations, denoted $v^{\mathrm{maj}}$,
\item the prior $p^{\mathrm{maj}}$ corresponding to $v^{\mathrm{maj}}$.
\end{itemize}
Examples of the most likely associations computed above are shown in the first
three rows of Table~\ref{tab:association-examples}. For the \verb|Niger-Congo|
language family, the most likely value for the feature \verb|Green_and_Blue|
(observed 9 times) is \verb|3 Black/green/blue|, with the corresponding prior
$0.667$. For the \verb|Bantoid| language genus, the feature
\verb|Green_and_Blue| was observed twice, both times with the same value
\verb|3 Black/green/blue|. The areal example corresponds to the neighborhood of
Yoruba, $(\phi,\lambda){=}(8.0,4.3)$ (where $\phi$ denotes latitude), for which
13 \verb|Green_and_Blue| features were observed, with the most likely value
corresponding to \verb|3 Black/green/blue| with prior $0.692$.

In addition for the implicational features, we stored the prior
probability of $v_i$ given the conditional feature $f_i$ from
equation~\ref{eqn:impl}, and the total count for $f_i$.  As an example
of a weak implicational preference consider the example given in the
fourth row of Table~\ref{tab:association-examples}, which means that
if a language has \verb|2 Red/yellow| for the feature
\verb|Red_and_Yellow|, then there is a slight preference ($p{=}0.583$,
estimated on the basis of 12 examples) for having
\verb|3 Black/green/blue| as the value for \verb|Green_and_Blue|. In
other words, \verb|2 Red/yellow| $\supset$ \verb|3 Black/green/blue|.
There were 54 cases of \verb|Green_and_Blue| in the training data, for
which the estimated a priori probability for \verb|3 Black/green/blue|
is $0.148$.

\begin{table}
  \centering%
  \includegraphics[width=\linewidth]{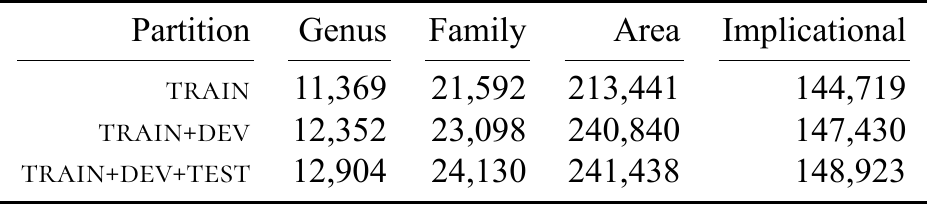}
  \caption{Precomputed typological feature associations.}
  \label{tab:associations}
\end{table}

Table~\ref{tab:associations} shows the overall sizes of the
association tables for the genetic, areal and implicational types
described above for the different partitions of the shared task
data.

\subsection{Sparse Language Vectors}
\label{subsec:sparse}
For each typological feature we train a separate feature estimator,
resulting in 185 estimators overall. When training an individual
estimator, we represent each language in the training and development
set as a sparse feature vector. Likewise, the languages whose features
need to be predicted at test time are also represented similarly.

\begin{table}
  \centering
  \includegraphics[width=\linewidth]{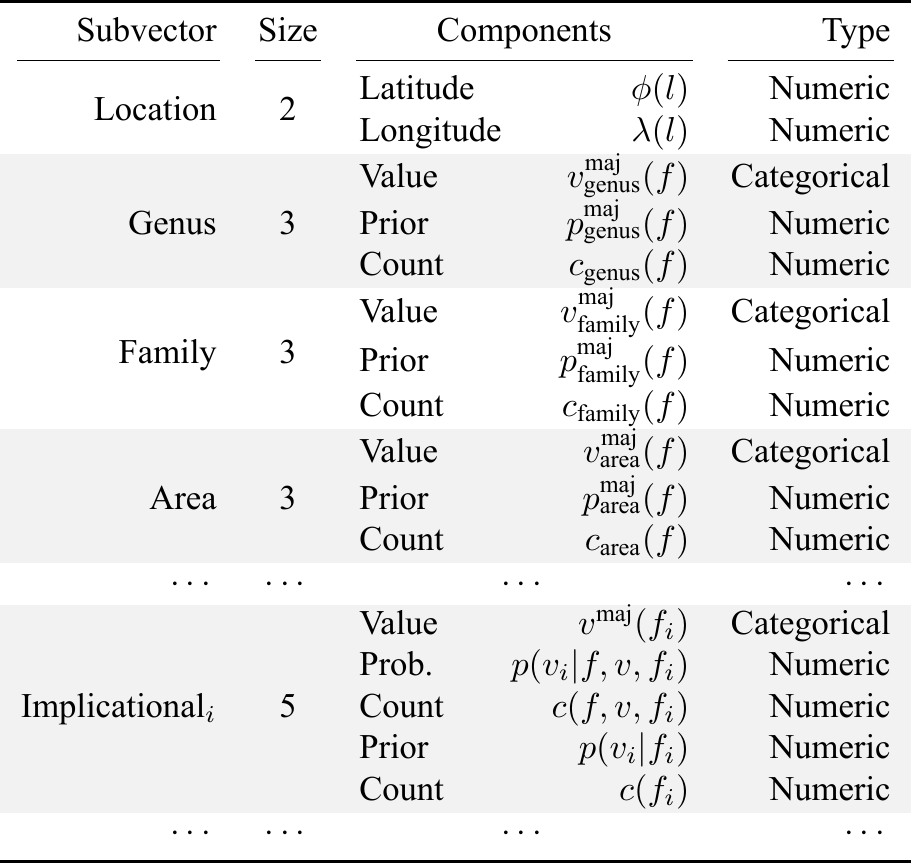}
  \caption{Language vector for typological feature $f$ and language $l$.}
  \label{tab:input-space}
  \vspace{-0.2cm}%
\end{table}

The makeup of individual language vector for a given typological
feature $f$ and a language $l$ is shown in
Table~\ref{tab:input-space}.  The vector consists of dense and sparse
subvectors; the components shown in the first four rows of the table
are mostly dense. The first subvector consists of language's latitude
$\phi$ and longitude $\lambda$ coordinate, represented as two numeric
features exactly the same as given in the shared task data. There is
no particular rationale for this choice other than we previously found
that for a different task (speech synthesis) choosing alternative
representations for the language location (e.g., distances to all
other languages in the training set) in the input features did not
significantly improve the results~\cite{gutkin2017}.

The next three subvectors representing genus, family and area, are
structured similarly using the three components used in association
tables described previously: the majority value $v^{\mathrm{maj}}(f)$
represented as a categorical feature, the prior corresponding to this
value $p^{\mathrm{maj}}(f)$ and the feature frequency $c(f)$, both
represented as numeric features. For these three subvectors, the
missing values are represented by the three-tuple $(v_{\emptyset},
10^{-6}, 0)$, where $v_{\emptyset}$ denotes a global dummy typological
feature value.\footnote{The choice of $10^{-6}$ for a missing value
prior is arbitrary. Any small non-zero value valid for a log
transform is suitable.}

The first four subvectors described above are followed by multiple
subvectors representing individual universal implications, as shown in
the fifth row of Table~\ref{tab:input-space}. Each implicational,
describing the dependence of feature $f_i$ on $f$, is represented as a
five-tuple whose elements are stored in the associations table for
implicational universals: The most likely value
$v^{\mathrm{maj}}(f_i)$ of $f_i$ corresponding to the highest
conditional probability $p(v_i | f, v, f_i)$ (interpreted as
probability of $f_i$ taking value $v_i$ given that $f$ is $v$), the
total count $c(f_i, f, v)$ of $f_i$ when $f$ is $v$, the prior $p(v_i|f_i)$
and the total count $c(f_i)$ for $f_i$ when $f$ is $v$.  The
missing implicational is represented as a five-tuple
$(v_{\emptyset},10^{-6}, 0, 10^{-6}, 0)$.  As mentioned above, the
implicational portion of the language vector is very sparse because,
for a feature $f \in F$ the language vector belongs to, one needs to
compute all its correlations to other features $f_i \in F$, where $F$
is the set of all 182 known features. Since the typological database
is very sparse, most of the observed correlations between $f$ and $f_i$
for a given language $l$ are poorly instantiated.

\begin{figure*}[ht]
  \centering%
  \includegraphics[trim={8.5cm 11.0cm 7.0cm 9.0cm},clip,width=\linewidth]{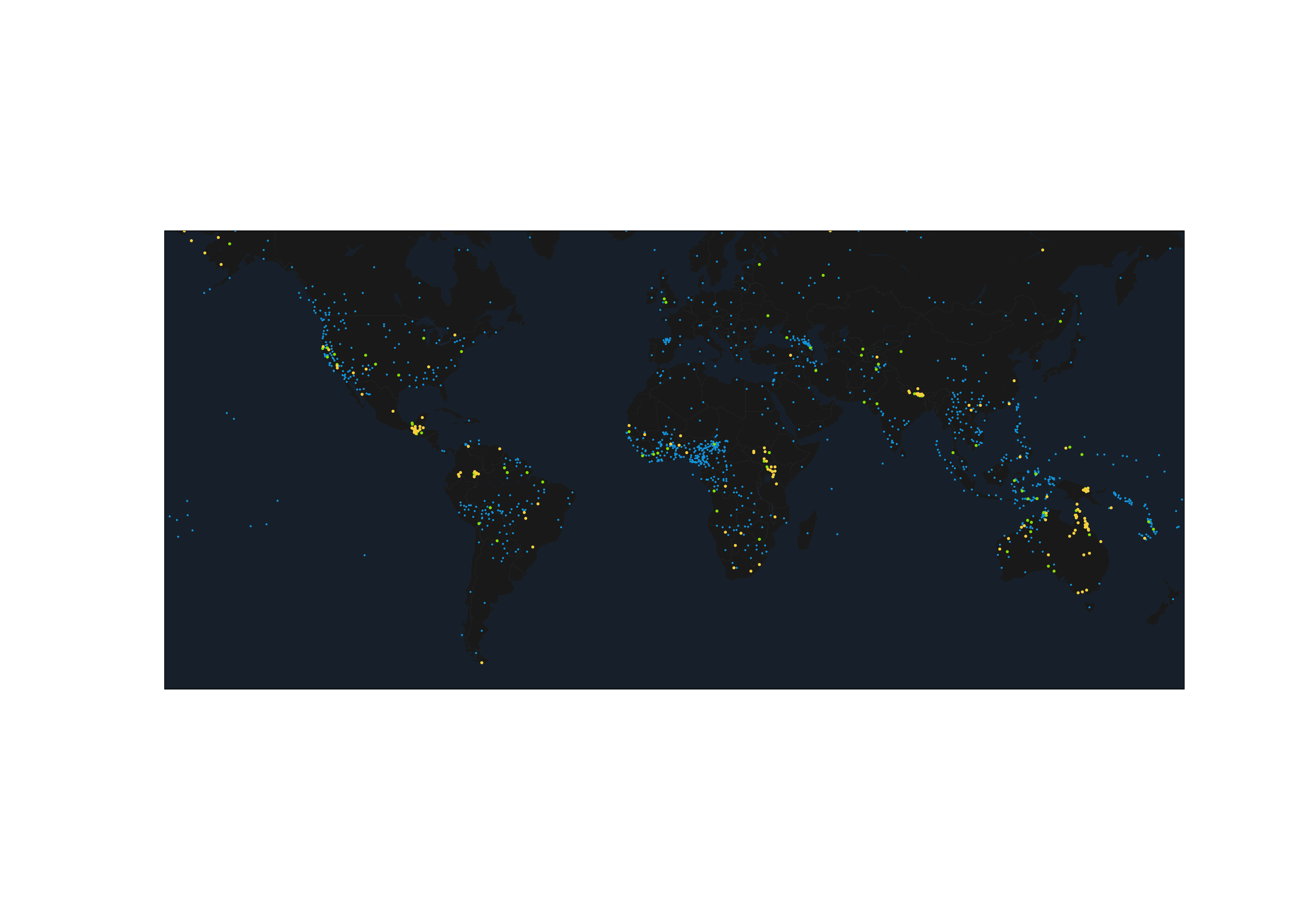}
  \caption{Languages in the \textsc{SIGTYP} training (blue), development (green) and test (yellow) sets.}
  \label{tab:world-map}
\end{figure*}

The categorical features in the language vector are represented using
a one-hot encoding and the numeric features are scaled to zero mean
and unit variance. For probability components, prior to numeric
feature scaling, the probability features are transformed into log
domain.

The overall representation results in the language vectors with a
rather high dimensionality. For example, the language vectors for the
\verb|Order_of_Subject,_Object_and_Verb| feature have the dimension of
4111.
As we shall see below from the shared task details, this
representation may already be too specific given the training set
which only contains 1,125 data points. This observation also explains
our choice of representing features and implicationals with the
attributes associated with their most likely majority values rather
than all the values for that feature observed in the data, as
suggested by a reviewer --- doing so will dramatically increase the
dimension of input feature space even further and render out approach
completely intractable.

\vspace{-0.2cm}%

\section{Experiments and Discussion}
\label{sec:experiments}
In what follows, we provide a brief overview of the datasets,
introduce the baseline systems we evaluated against during the
development of our method, provide the evaluation results of
miscellaneous machine learning algorithms on the development set that
guided our final model selection, describe the two configurations
submitted to the constrained subtask and, finally, mention the
approaches that did not work well in our settings.

\subsection{Data Overview}
\begin{table}
  \centering%
  \includegraphics[width=\linewidth]{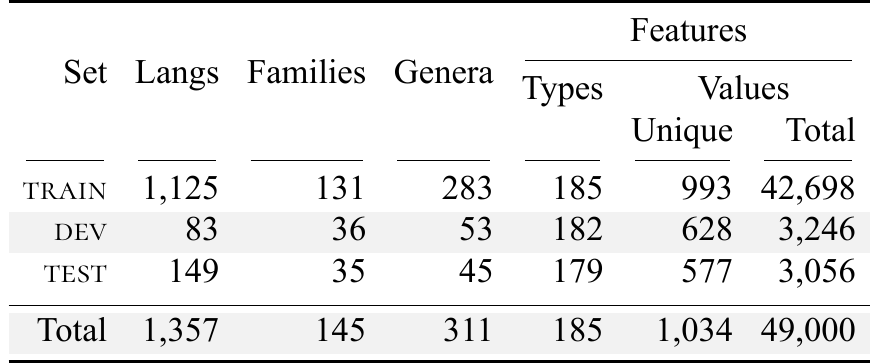}
  \caption{The three partitions of \textsc{SIGTYP} data.}
  \label{tab:corpus-stats}
  \vspace{-0.2cm}%
\end{table}
Some of the properties of the data used in our
experiments\footnote{The \textsc{SIGTYP} data is available from
  \url{https://github.com/sigtyp/ST2020/tree/master/data}.}  are shown
in Table~\ref{tab:corpus-stats}, where we summarize the counts for
three partitions of the data: the training and development sets, and
the \emph{blind} test set. For each of the sets, the total number of
languages is shown along with the number of unique language families
and genera. The count of typological feature types is displayed along
with the number of unique feature values and the total number of
observed values. As can be seen from the table, the size of the data
is very small, which precludes us from using state-of-the-art deep
learning methods, especially given our approach to representing each
language as a point in high-dimensional space.

The 1,357 longitude and latitude language coordinates provided in the
data are displayed in Figure~\ref{tab:world-map} (best viewed in
color) using the Mercator projection. The languages in the training
set are shown in blue, the development set languages in green and
the test set languages in yellow.

\subsection{Baselines}
\newcommand{\base}[1]{\textsc{b}$_{#1}$}
For our baselines five simple prediction algorithms based on
deterministic search were implemented. Initially we evaluated these
systems on the development (\textsc{dev}) set and later on the test
set, after the golden truth data was released by the organizers.

The \emph{global} majority class predictor (denoted \base{1})
accumulates the frequencies of all the typological feature values in
the training data and predicts the most frequent value for the feature
in question irrespective of its phylogenetic or areal attributes. The
\emph{clade} majority class predictor (denoted \base{2}) extends the
previous predictor by also taking into
account language genera and families, producing predictions of the form
\begin{displaymath}
  \hat{v}_{\text{\textsc{B}}_2}(f) = \begin{cases}
    v^{\mathrm{maj}}(f) & \splitcond{\text{if $c_{\mathrm{genus}}(f)=0$ and} \\
      \;\;\;$c_{\mathrm{family}}(f)=0$ \, ,} \\
    v^{\mathrm{maj}}_{\mathrm{family}}(f) & \text{if $c_{\mathrm{genus}}(f)=0$} \, , \\
    v^{\mathrm{maj}}_{\mathrm{genus}}(f) & \text{if $c_{\mathrm{genus}}(f)>0$} \, .
  \end{cases}
\end{displaymath}
Predictor \base{3} only relies on the distances between languages'
geographic coordinates defined in the data. For each typological
feature in question, the predictor returns the value belonging to the
closest language, according to the Haversine formula~\cite{robusto1957},
with a known value for that feature. The Haversine distance
$d(l_i,l_j)$ is computed between each pair of languages $l_i$ and
$l_j$ represented as points on an ideal sphere using their respective
latitude and longitude coordinates $(\phi_i, \lambda_i)$ and $(\phi_j,
\lambda_j)$.

A na{\"\i}ve approach for combining areal and phylogenetic knowledge
is implemented by predictor \base{4} which performs its search in two
additional clusters of inter-language distances, grouped by genera and
families: The algorithm first searches for the known feature within
the geographically closest languages of the same genus, followed by a
search within the closest languages from the same family, falling back
to the closest languages from the global set.

The final baseline configuration \base{5} uses an ensemble approach by
combining areal and phylogenetic information by using majority
voting. The areal estimate is provided by the majority estimate from
the candidate's neighborhood (as described in
Section~\ref{subsec:precomp}) or, if that information is not available
due to the nearest language being located beyond the neighborhood
radius, by the known feature from the closest languages outside the
neighborhood. The phylogenetic clade estimates are provided by the
majority class values from genera and families (\base{2}),
respectively. The final estimate is produced by the majority voting,
where at least two predictions have to agree, otherwise the predictor
falls back to global majority class estimates (\base{1}).

\begin{table}
  \centering%
  \includegraphics[width=\linewidth]{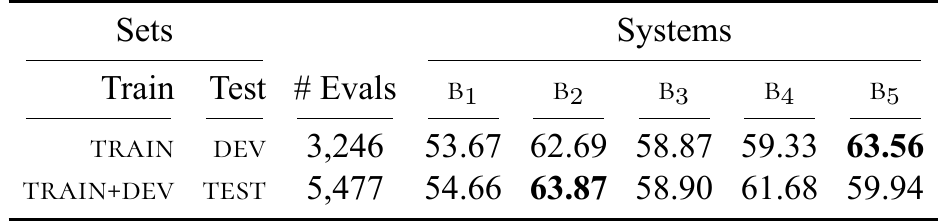}
  \caption{Micro-averaged accuracies (\%) for the five baseline systems.}
  \label{tab:baselines}
  \vspace{-0.2cm}%
\end{table}
The micro-averaged accuracies for five of the baseline systems are shown
in Table~\ref{tab:baselines}. We performed the evaluation by first
obtaining the estimates from a set shown in the first column and then
predicting \emph{each} known feature value for the languages from a
set shown in the second column (\textsc{dev} or golden \textsc{test})
by pretending that this value is unknown. Best accuracies are
highlighted in bold. As can be seen from the table, the best baseline
configuration on the \textsc{dev} set is the \base{5} ensemble method,
with the pure clade-based \base{2} coming second. On the golden
\textsc{test} set, \base{2} is the winning configuration, while the
second best configuration is \base{4}. It is interesting to note, that
the methods that completely ignore phylogeny (\base{1} and \base{3})
did not perform well in either evaluation.

As one reviewer notes, the apparent predominance of phylogeny in our
results may seem surprising given that areal features are well-known
to be important in many areas of the world --- e.g. India
\citep{emeneau56}. This is likely due at least in part to the fact
that many of the language families in the sample are small families
spoken in a relatively limited geographic area, or else are families
like Pama-Nyungan, which are more or less isolated from unrelated
neighbors.  This would tend to confound the influence of phylogeny
versus geography, since it is only when \emph{part} of a language
family is spoken in an area that is populated by speakers of unrelated
families that one will see robust effects of geography.

\subsection{Model Selection Using Development Set}
\label{subsec:model-selection}%
\begin{table}
  \centering%
  \includegraphics[width=\linewidth]{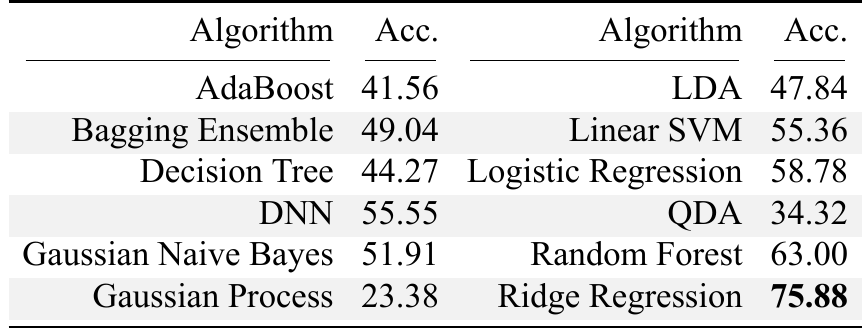}
  \caption{Micro-averaged accuracies (\%) of various algorithms on the
    development set.}
  \label{tab:dev-algos}
  \vspace{-0.2cm}%
\end{table}
We approach the feature prediction task by training 185 multi-class
classifiers, one for each typological feature. Several standard
machine learning algorithms were evaluated on the development set to
establish the most optimal algorithm for the task. In particular, we
trained decision trees~\cite{breiman1984}, Naive
Bayes~\cite{langley1992} with Gaussian mixtures, Gaussian
processes~\cite{rasmussen2006}, classifiers based on linear and
quadratic discriminant analysis (LDA and QDA,
respectively)~\cite{tharwat2016}, support vector machines (SVM) with
linear kernel~\cite{suykens1999}, multinomial logistic
regression~\cite{bohning1992}, ridge regression~\cite{hoerl1970} and
simple feed-forward neural networks with a single layer of 200 units
(DNN). In addition, three ensemble configurations were also evaluated:
multi-class AdaBoost (with 100 estimators)~\cite{zhu2009}, random
forests (with 200 estimators, minimum of three samples per leaf and
information gain as splitting criterion)~\cite{breiman2001} and
bagging ensembles~\cite{breiman1999}. For all the algorithms we used
the implementation provided by \texttt{scikit-learn}
toolkit.\footnote{\url{https://scikit-learn.org/}} We mostly used
default hyper-parameters provided performing no special tuning.

The evaluation results for each of the classifiers are shown in
Table~\ref{tab:dev-algos}. As can be seen from the table, there are
only five algorithms out of twelve which are remotely comparable to
our development set baselines summarized in Table~\ref{tab:baselines}:
DNN, linear SVM, multinomial logistic regression, random forest and
ridge regression. Out of those, the random forest ensemble
method and the ridge regression are the only competitive models, with
ridge regression estimation strongly outperforming all of our
baselines on the development set achieving the micro-averaged accuracy
of 75.88\%. On the basis of this result we chose ridge regression
algorithm for our experiments on the test set and the final
submission.

It is worth noting, however, that it is not difficult to find cases
where the ridge regression estimator does not perform as well as other
methods. For example, for feature \verb|SVNegO_Order|, the DNN (93\%
accuracy) and SVM (87\% accuracy) both outperform ridge regression
(80\%). We hypothesized that a better alternative to fixing a certain
method for all the features is to employ an ensemble of classifiers
which are feature-specific. Although we include this approach in our
implementation, so far we only tested it under scenario of performing
an iterative stratified $k$-fold cross-validation~\cite{witten2005}
over the training, rather than development, set. The optimal
feature-specific classifiers obtained by this method did not fare well
on the development set, which is probably an indicator that
development set was not an optimal reflection of the training data.

Aside from the ridge regression and random forest ensemble methods, the
poor average performance of other classification algorithms on the
development set is rather unexpected. We hypothesize that this may be
due to several confounding factors. The first issue is that our
approach of representing training data for each typological feature as
sparse language vectors results in small amounts of high-dimensional
training data that may not be enough to reliably train most of the
classifiers in our particular setting. The second issue is the high
feature sparsity which adversely affects multi-class classification
for most of the algorithms. Consider the
\verb|Order_of_Subject,_Object_and_Verb| feature. The frequencies of
its seven values (corresponding to class label counts) in the training
data represented as an ordered set are $\{311, 226, 113, 51, 16, 6,
1\}$. Although we employ balanced class weighting for all the
algorithms, using the label values to adjust the weights inversely
proportional to class frequencies observed in the data, this may not
be enough. Some unsatisfactory attempts to further remedy this are
described below.

We further note that the relatively good performance of ridge
regression and random forest classifiers on the development data
without any hyper-parameter fine-tuning may possibly be explained by
the relative robustness of both methods to sparseness and collinearity
effects (which severely affect other types of parametric and
non-parametric predictors in our experiment), as previously analyzed
in detail by~\citet{tomaschek2018} and observed by others in a
typological setting~\cite{burdick2020} and
elsewhere~\cite{josifoski:2019}. We briefly reevaluate these
assumptions on the released test set in Section~\ref{subsec:disc}.

\subsection{Shared Task Submission}
Our submission to shared task consists of two ridge regression
classifier configurations denoted \texttt{NEMO\_system1} and
\texttt{NEMO\_system2}.\footnote{Submitted as
{\scriptsize\texttt{NEMO\_system1\_assoc-train-dev\_constrained}} and
{\scriptsize\texttt{NEMO\_system2\_assoc-train-dev-test\_constrained}},
respectively.} The difference between the two configurations is in how
the training data is generated from the original shared task
subsets. The training data for \texttt{NEMO\_system1} classifier
consists of the language vectors generated from the original training
(\texttt{train.csv}) and development (\texttt{dev.csv}) sets only.
For \texttt{NEMO\_system2}, the training data also includes the
phylogenetic, areal and implicational relations computed from the
\emph{known} features in the blind test set
(\texttt{test\_blinded.csv}).

\subsection{Approaches That Disappointed}
In addition to the experiments described above, we also tried applying
missing feature imputation algorithms to the data provided, as a
preprocessing step prior to training the classifiers described
above. Neither of the feature imputation algorithms that we tried
produced satisfactory results that could improve upon our best baseline.
One of the imputation approaches that we evaluated was Multiple
Imputation with Denoising Autoencoders (MIDAS) by~\citet{lall2020}.
This approach is particularly attractive because it natively supports
categorical features. However, due to reliance on a denoising
autoencoder architecture~\cite{vincent2010}, the network requires
large amounts of training data to be estimated reliably, which
prevented us from getting adequate performance on the provided
typological data.\footnote{The MIDAS imputer for linguistic
  typological data is released with the rest of our software.}

Similarly disappointing were the attempts to address the class
imbalance problem during multi-class classifier training using class
resampling techniques. Similar to the feature imputation above,
resampling was applied as a preprocessing step prior to training the
classifiers. In particular, neither the application of synthetic
minority over-sampling technique (SMOTE)~\cite{chawla2002} nor the
adaptive synthetic sampling~\cite{he2008} produced any improvements
over our best baseline.

We also extended our method to use country code information provided
with the task data for each language. This was achieved by
accumulating the per-country typological feature priors, similar to
other areal and phylogenetic features, and representing the country
codes as categorical features in classification, the process described
in Section~\ref{sec:method}. Ridge regression estimator that included
country code information achieved micro-averaged accuracy of 75.66\%
on the development set -- a small deterioration of 0.3\% compared to
our best result. This may be due to multiple errors in the provided
data, e.g., the country code for the Chepang language from
Sino-Tibetan family spoken in Nepal~\cite{caughley1982} is defined as
\texttt{US} in the development set.

\subsection{Discussion}
\label{subsec:disc}

\begin{table}
\centering%
\footnotesize%
\begin{tabular}{lrrr}
\toprule%
\multirow{2}{*}{\textbf{Family}} & \multicolumn{2}{c}{\textbf{\# Languages}} & \multirow{2}{*}{\textbf{Acc.}}\\
& \textbf{Train} & \textbf{Test} & \\
\cmidrule(lr){1-1}\cmidrule(lr){2-2}\cmidrule(lr){3-3}\cmidrule(lr){4-4}
Mayan  & 6 & 17 & 0.73\\
\rowcolor{light-gray}
Mahakiranti & 4 & 13 & 0.72\\
Tucanoan & 3 & 8 & 0.67\\
\rowcolor{light-gray}
Nilotic & 3 & 15 & 0.65\\
Madang & 3 & 9 & 0.68\\
\rowcolor{light-gray}
N. Pama-Nyungan & 2 & 24 & 0.67\\
\bottomrule%
\end{tabular}
\normalsize%
\caption{\label{tab:families} Numbers of languages from each test family in the
  test and training data, along with our best system's per-family accuracy.}
\vspace{-0.4cm}%
\end{table}
\begin{figure}[t]
  \centering%
  \includegraphics[width=\linewidth]{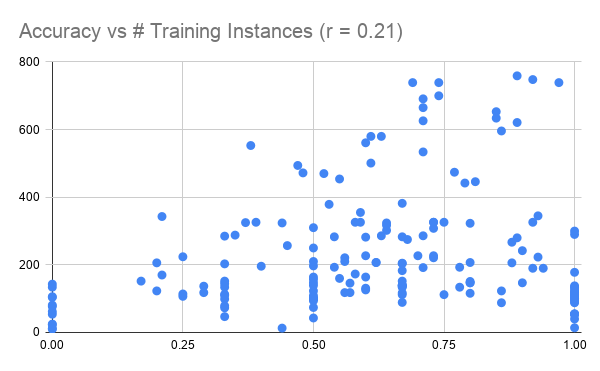}
  \caption{Feature accuracy versus number of instances of the feature in the
    training and development data ($r{=}0.21$).}
  \label{fig:feat-acc}
  \vspace{-0.4cm}%
\end{figure}
\begin{table*}
  \centering%
  \adjustbox{width=\textwidth}{
\begin{tabular}{lrrr}
\textbf{Feature} & \textbf{baseline} & \textbf{our best} & \textbf{\# in test}\\
\verb|Inclusive/Exclusive_Distinction_in_Verbal_Inflection| & 0.00 & 0.75 & 4\\
\verb|Rel_between_the_Order_of_Obj_and_Vb_and_the_Order_of_Adj_and_Noun|
  & 0.14 & 0.86 & 35\\
\verb|Order_of_Object_and_Verb| & 0.26 & 0.89 & 38\\
\verb|Order_of_Adposition_and_Noun_Phrase| & 0.26 & 0.89 & 27\\
\verb|Epistemic_Possibility| & 0.20 & 0.80 & 5\\
\end{tabular}
}
\caption{\label{tab:best} Five features on which our best system most
  outperformed the best baseline, with accuracies for each, and counts for each
  in the test data.}
\end{table*}
\begin{table*}
\centering%
\footnotesize%
\begin{tabular}{lrrr}
\textbf{Feature} & \textbf{baseline} & \textbf{our best}  & \textbf{\# in test}\\
\verb|Person_Marking_on_Adpositions| & 0.50 & 0.20 & 10\\
\verb|Genitives,_Adjectives_and_Relative_Clauses| & 1.00 & 0.67 & 3\\
\verb|Noun_Phrase_Conjunction| & 0.67 & 0.33 & 3\\
\verb|Subtypes_of_Asymmetric_Standard_Negation| & 1.00 & 0.66 & 5\\
\verb|Obligatory_Possessive_Inflection| & 1.00 & 0.50& 6\\
\end{tabular}
\normalsize%
\caption{\label{tab:worst} Five features on which our best system most
  underperformed the best baseline, with accuracies for each, and counts for each
  in the test data.}
\end{table*}
We start with some observations about the rather apparent differences between
the test data and the training and development data. In Table~\ref{tab:families}
we list the six language families evidenced in the test data, and the numbers of
languages for each represented in the training and development data, versus the
test data. As can be seen some of the language families are significantly more
represented in the test data than they are in the data that was released
earlier. This is particularly the case for Northern Pama-Nyungan, which
comprised two languages in the training-development data, and twenty four
languages in the test data. This skew can also be seen in
Figure~\ref{tab:world-map} where the yellow dots represent the test data. There
are islands of yellow in Australia, Northern South Asia, Central East
Africa and in Central America, surrounded by seas of green and blue.\footnote{
As one of the reviewers notes, it would be useful for future tasks if the
organizers could provide a rationale for the data splits chosen.}

Turning now to our results, we discuss here the performance of our
best model \verb|NEMO_system2|
(overall micro-averaged accuracy $0.66$).  First of all, we note that
there is very little correlation between the number of exemplars of a
feature in training and development, and our system's performance on
that feature in the test set (correlation coefficient $r{=}0.21$): see
Figure~\ref{fig:feat-acc}.\footnote{A reviewer asks if there is any correlation between accuracy and the number of value settings for each feature in the training and development set. This correlation is somewhat higher, and negative ($r{=}-0.34$), not surprising since we would expect to do worse if there are more possible values.}

Second, comparing our results against the best baseline provided by
the task organizers --- \verb|frequency-baseline_constrained| (overall
micro-averaged accuracy $0.51$), we list in Tables~\ref{tab:best} and
\ref{tab:worst} the five features for which our system had the largest
win and loss, respectively in terms of absolute accuracy difference.
Note that three of the five features for which we showed the largest
gains relate to word order. While this may be due to the automatically
derived implicational features used in our models ---
word-order-related features being probably among the most robust of
the implicational universals --- it is also true that these features
are better instantiated in the data.

\begin{table}
  \centering%
  \includegraphics[width=\linewidth]{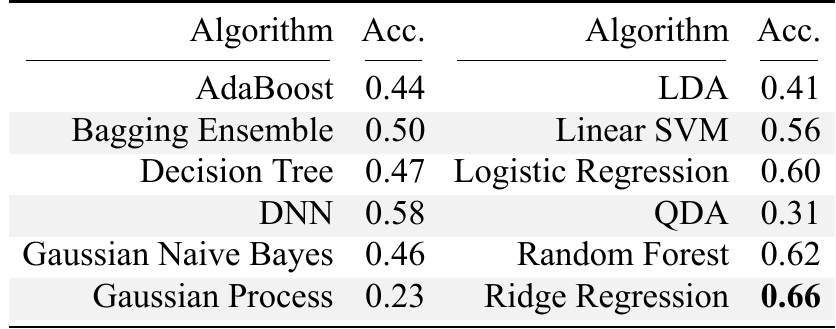}
  \caption{Micro-averaged accuracies of various algorithms on the
    test set.}
  \label{tab:test-algos}
\end{table}

One reviewer notes that it would be interesting to see the results of
ablation studies for some of the high dimensional features discussed
in Section~\ref{subsec:sparse}. We agree, but leave this for future
work.

Finally, recall from the discussion of development set-based model
selection in Section~\ref{subsec:model-selection} that ridge
regression was the best performing model, followed by the
non-parametric random forest predictor (see
Table~\ref{tab:dev-algos}). Once the golden test data was released, we
evaluated the same set of twelve classifiers on the test set with the
results shown in Table~\ref{tab:test-algos}. As can be seen from the
table, the ranking of three best classifiers on the development set is
preserved on the held-out test set as well: ridge regression is the
best-performing classifier in our task, followed by random forest and,
finally, multinomial logistic regression.

\section{Conclusion}
We have presented the NEMO submission to the 2020 \textsc{SIGTYP}
shared task. Our system used genetic, geographical and automatically
deduced implicational universals, and a range of classifiers of which
ridge regression yielded the best overall performance. Our method
achieved 0.66 overall accuracy on the task, compared to the best
baseline provided by the organizers, which had an accuracy of
0.51. Our system tended to do better on test language families that
were better represented in the training and development data, though
interestingly the correlation between the number of instances
of \emph{features} in the training/development data and performance on
that feature was not strong.

As discussed above, we include in the training package methods for ensembling
classifiers so that one can find per-feature optimal classifiers, but we did not
make use of this functionality for our submitted results. We do think, however,
that further experimentation with this functionality would be useful.

Finally, as one of the reviewers notes, it would be interesting to ask what use work along the lines presented here could be to the field linguist/typologist who is interested in testing potential relationships between languages based on shared features.  Many of the systems reported in Section~\ref{sec:related} had this aim in mind.
While we do not want to overstress the value of this sort of work compared to good linguistic intuition, we do think that knowing the Bayesian priors of associations, as we discussed in Section~\ref{subsec:precomp} could at least serve as a reminder that some feature settings may be shared simply because they are very common anyway.

\section*{Acknowledgments}
The authors thank the anonymous reviewers for many helpful suggestions on an
earlier version of this paper.

\bibliographystyle{acl_natbib}
\bibliography{anthology,main}

\end{document}